\documentclass{article}

\usepackage{PRIMEarxiv}

\usepackage[utf8]{inputenc} 
\usepackage[T1]{fontenc}    
\usepackage{cite}
\usepackage{amsmath,amssymb,amsfonts}
\usepackage{algorithmic}
\usepackage{graphicx}
\usepackage{textcomp}
\usepackage{siunitx}
\usepackage{caption}
\usepackage{subcaption}
\usepackage{booktabs}
\usepackage{multirow}
\usepackage{multicol}
\usepackage{wrapfig}
\usepackage{balance}
\usepackage[square, numbers, sort&compress]{natbib}
\newcommand{\etal}{\textit{et~al.}}
\graphicspath{{media/}}     

\title{Self-Supervised Pretraining Improves Performance and Inference Efficiency in Multiple Lung Ultrasound Interpretation Tasks
}

\author{
  Blake VanBerlo\thanks{B. VanBerlo is a Vanier Scholar (FRN 186945) supported by the Natural Sciences and Engineering Research Council of Canada (NSERC).}  \\
  Cheriton School of Computer Science \\
  University of Waterloo \\
  Waterloo, Canada \\
  \texttt{bvanberl@uwaterloo.ca} \\
   \And
  Brian Li \\
  Department of Systems Design Engineering \\
  University of Waterloo \\
  Waterloo, Canada \\
    \And
  Jesse Hoey \\
  Cheriton School of Computer Science \\
  University of Waterloo \\
  Waterloo, Canada \\
    \And
  Alexander Wong \\
  Department of Systems Design Engineering \\
  University of Waterloo \\
  Waterloo, Canada \\
}

\begin{document}
\maketitle

\begin{abstract}
In this study, we investigated whether self-supervised pretraining could produce a neural network feature extractor applicable to multiple classification tasks in B-mode lung ultrasound analysis.
When fine-tuning on three lung ultrasound tasks, pretrained models resulted in an improvement of the average across-task area under the receiver operating curve (AUC) by $0.032$ and $0.061$ on local and external test sets respectively.
Compact nonlinear classifiers trained on features outputted by a single pretrained model did not improve performance across all tasks; however, they did reduce inference time by $49\%$ compared to serial execution of separate fine-tuned models.
When training using $1\%$ of the available labels, pretrained models consistently outperformed fully supervised models, with a maximum observed test AUC increase of $0.396$ for the task of view classification.
Overall, the results indicate that self-supervised pretraining is useful for producing initial weights for lung ultrasound classifiers.
\end{abstract}

\keywords{Multi-task \and self-supervised learning \and ultrasound}

\section{Introduction}
\label{sec:introduction}

\begin{figure*}[h]
    \centering
    \includegraphics[width=\textwidth]{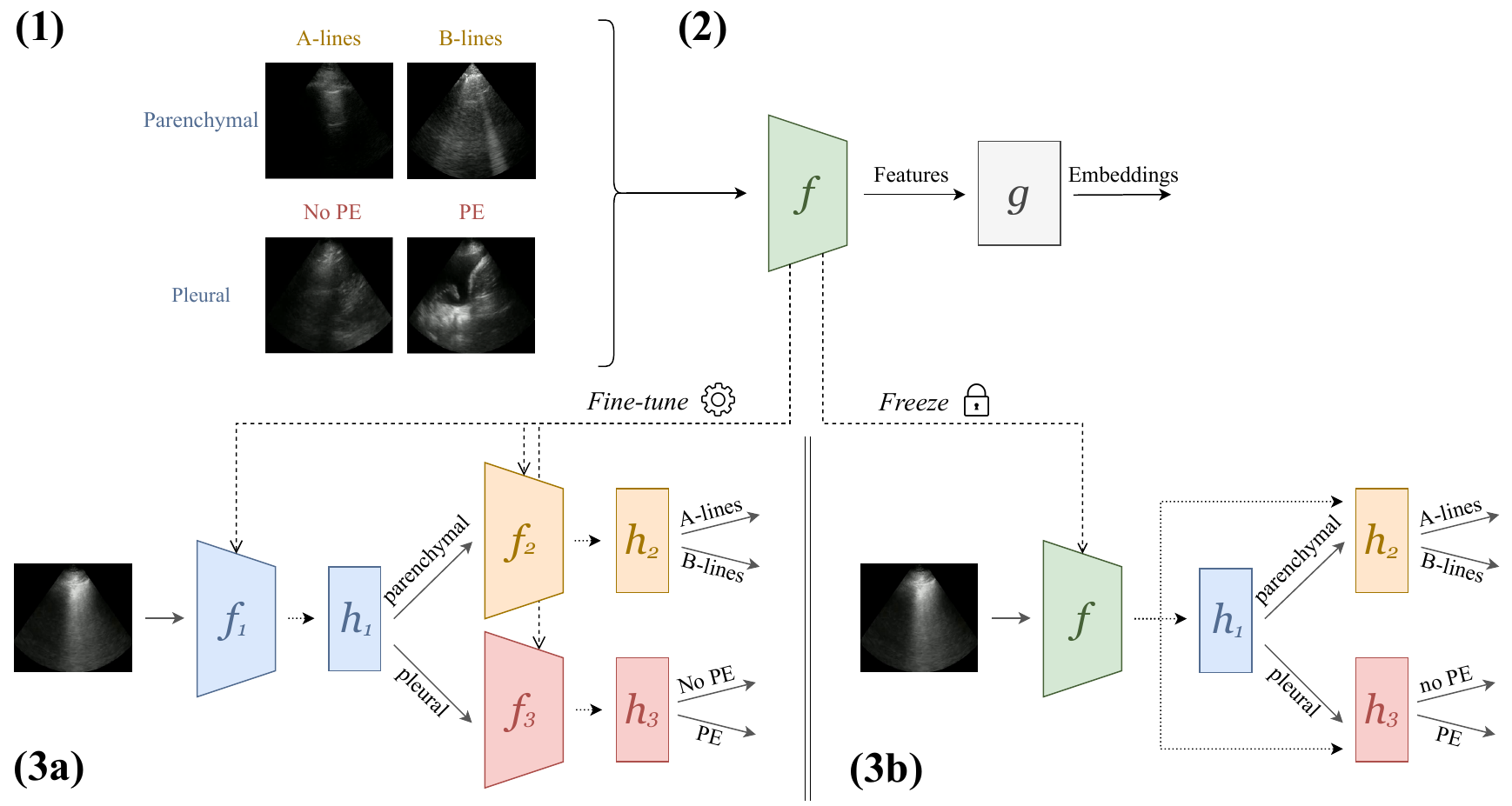}
    \caption{An overview of the methods described in this work. \textbf{(1)} Three tasks were identified for lung ultrasound (LUS) image classification: parenchymal versus pleural views, A-lines versus B-lines (applicable to parenchymal views), and pleural effusion (PE) versus no pleural effusion (applicable to pleural views). \textbf{(2)} A convolutional feature extractor $f$ was pretrained to minimize a self-supervised objective, using unlabelled and labelled LUS images as input and trainable projector $g$. \textbf{(3a)} Task-specific models were defined by appending linear classifier or multilayer perceptron $h_i$ to copies of pretrained $f$. The models were trained end-to-end for each task using labelled data. \textbf{(3b)} An alternative framework in which $f$'s weights were not fine-tuned. Instead, task-specific models $h_i$ were trained that each received $f$'s features as input. }
    \label{fig:overview}
\end{figure*}

Recent years have witnessed a surge of interest in self-supervised learning (SSL) as a strategy for representation learning in computer vision. 
Hailed as a means to productively leverage unlabelled data when labels are scarce, self-supervised pretraining has been shown to improve performance on several supervised learning tasks in multiple domains of medical imaging, such as radiography~\cite{azizi_big_2021, haghighi_dira_2022}, computed tomography~\cite{zhou_models_2021, haghighi_dira_2022}, magnetic resonance imaging~\cite{zhou_models_2021, haghighi_dira_2022}, ultrasound~\cite{perek_self_2021,basu_unsupervised_2022}, and dermatology~\cite{azizi_big_2021}. 
Self-supervised pretraining produces a feature extractor that may be used to initialize the weights of a model in a supervised learning setting. 
Studies have indicated that models pretrained with SSL perform comparably to fully supervised models even when fine-tuned with significantly less labelled data~\cite{azizi_big_2021, zhou_models_2021}. 
Given the widespread paucity and expense of labelled medical images, it is therefore unsurprising that SSL has risen as a reasonable strategy to leverage unlabelled data.

Interpretation of medical images consists of completing several recognition tasks, occasionally in a hierarchical manner. 
In the hierarchical setting, interpreters engage in the predictive process of a decision tree, beginning with the root node and traversing down a single path, guided by decisions at each node. 
Examples include the distinction of malignant pulmonary nodules on CT~\cite{ma2008classification} and the identification of lipomas and liposarcomas on MRI~\cite{shim2020mri}. 
In this study, we focus on lung ultrasound (LUS) -- an examination involving the recognition of multiple artefacts that narrow differential diagnoses in emergency and critical care scenarios, hereafter referred to as \textit{multi-task LUS interpretation}.

Past work in machine learning-based hierarchical medical imaging classification has resorted to training entirely separate classifiers for each node in the tree~\cite{yoo2020deep}. 
This study sought to determine if a single feature extractor can produce meaningful representations for multi-task LUS interpretation. 
We hypothesized that self-supervised pretraining is suited for the task of developing a feature extractor that is useful for multiple classification tasks. 
Self-supervised pretraining produces feature representations that may be adapted for training multiple supervised learners, while making use of unlabelled examples. 
The feature extractor can be fine-tuned for individual subsequent tasks (Figure~\ref{fig:overview}, 3a).
Alternatively, the weights of the feature extractor may be held constant, facilitating the addition of new tasks to the multi-task LUS interpretion system by training nonlinearities on top of the features (Figure~\ref{fig:overview}, 3b). 
The contributions of this work are thus as follows: (1) an investigation of the suitability of self-supervised feature extractors for multi-task interpretation of B-mode LUS, and (2) a tree-based classification strategy in which the inputs to the root node are obtained from a feature extractor pretrained with SSL. 
The evaluation provides performance and runtime metrics for each task, comparing the fine-tuning of end-to-end models for each task against training a multilayer perceptron (MLP) on features yielded from one pretrained extractor. 

\section{Related Works}

\subsection{Joint Embedding Self-Supervised Learning}

Broadly, self-supervised learning (SSL) is a form of unsupervised representation learning that is employed to pretrain a feature extractor for transfer learning.
It consists of learning to solve a \textit{pretext task}, which is a supervised learning problem for which labels are computed from unlabelled data. 
The weights of the feature extractor are used to initialize a new model trained to solve a supervised learning task for which labels are available.
In the joint embedding framework of SSL, the pretext task is designed to reduce the differences between features of semantically related images that satisfy a pairwise relationship. 
Semantically related pairs of images are customarily passed through the feature extractor, with the output being sent through a projection head (typically a MLP), producing embeddings.
Contrastive learning methods seek to reduce the distance between embeddings of paired images (positive pairs) and increase the distance between embeddings of images that do not satisfy the pairwise relationship (negative pairs)~\cite{chen2020simple}.
Non-contrastive learning methods dispense with negative pairs, focusing only on minimizing distances between the embeddings of positive pairs~\cite{zbontar2021barlow,bardes2022vicreg}.

\subsection{Joint Embedding Methods in B-Mode Ultrasound}

Multiple studies have assessed the impact of Joint Embedding approaches to self-supervised pretraining on the performance of machine learning solutions in diagnostic B-mode US tasks, particularly when labels are scarce. 
Contrastive and non-contrastive methods have been applied to breast tumour classification and left ventricle segmentation, with mixed results~\cite{perek_self_2021,nguyen_semi-supervised_2021,saeed_contrastive_2022}.
Chen~\etal~\cite{chen2021uscl} proposed a custom contrastive learning objective with interpolated intra-video positive pairs, outperforming both fully supervised and SimCLR-pretrained models on the public POCOVID-Net dataset~\cite{born2020pocovid}. 
Adopting a curriculum learning approach, Basu~\etal~\cite{basu_unsupervised_2022} achieved even better performance on POCOVID-Net with their contrastive learning method that employed progressively harder intra-video positive pairs.

\subsection{Multi-Task Medical Image Interpretation}

Several studies have addressed multi-task learning for multi-task medical imaging interpretation. 
For instance, Zhang~\etal~\cite{zhang2021multi} trained a single neural network with dedicated output layers for the classification of carotid plaques and estimation of the degree of stenosis on CT angiography imaging.
Xu~\etal~\cite{xu2018less} proposed a single convolutional neural network (CNN) architecture for adbominal US view classification and landmark localization, using features from intermediate residual blocks as input for both tasks.
Focusing instead on hierarchical interpretation, Fu~\etal~\cite{fu2021mendt} proposed a system for medical image classification consisting of a convolutional neural network (CNN) followed by a decision tree in which each node is a linear classifier~\cite{fu2021mendt}.
Decision trees with neural network nodes have also been proposed~\cite{yoo2020deep}. 

In our study, we show that a single CNN pretrained with self supervision provides sufficient features for multiple tasks, including tasks arranged hierarchically. 
Note that the present work is distinct from multi-task learning in that it explores the utility of reusing a single self-supervised pretrained feature extractor for the development of multiple LUS classifiers.

\section{Methods}

\subsection{LUS Classification Tasks}

The LUS interpretative workflow addressed in this work has been described as a decision tree~\cite{soni2019point}. 
After determining the view, the interpreter traverses down the tree to look for increasingly specific artefacts that reduce a possible differential diagnosis.
We focus on three binary classification tasks for LUS image interpretation: view classification ({\tt View}), A-line versus B-line classification ({\tt A/B}), and pleural effusion detection ({\tt PE}). 
The former is applicable to parenchymal LUS views, and the latter to pleural LUS views. 
Table~\ref{tab:lus-tasks} summarizes these tasks, and Figure~\ref{fig:overview} displays emblematic examples for each class.

\begin{figure}
    \centering
    \begin{subfigure}[b]{0.31\linewidth}
         \centering
         \includegraphics[width=0.75\textwidth]{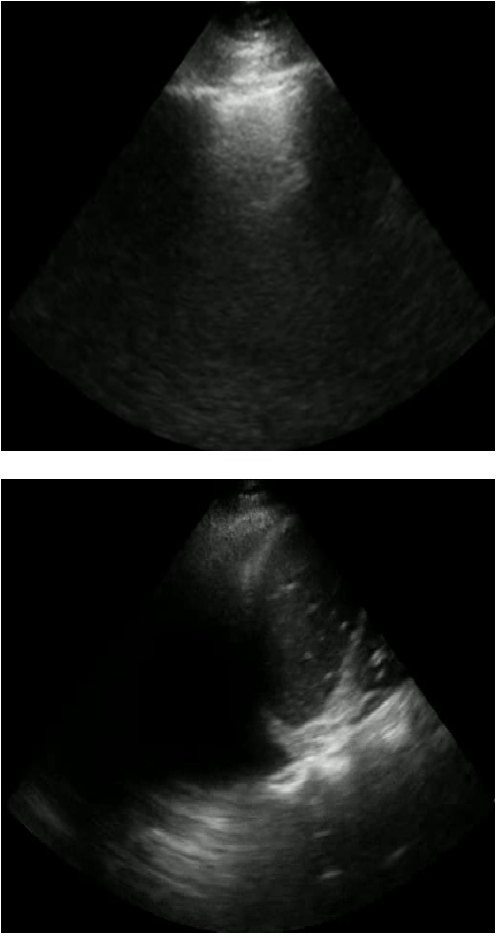}
         \caption{{\tt View}: Parenchymal~(top), pleural (bottom)}
         \label{fig:ft-tree}
     \end{subfigure}
     \hfill
     \begin{subfigure}[b]{0.31\linewidth}
         \centering
         \includegraphics[width=0.75\textwidth]{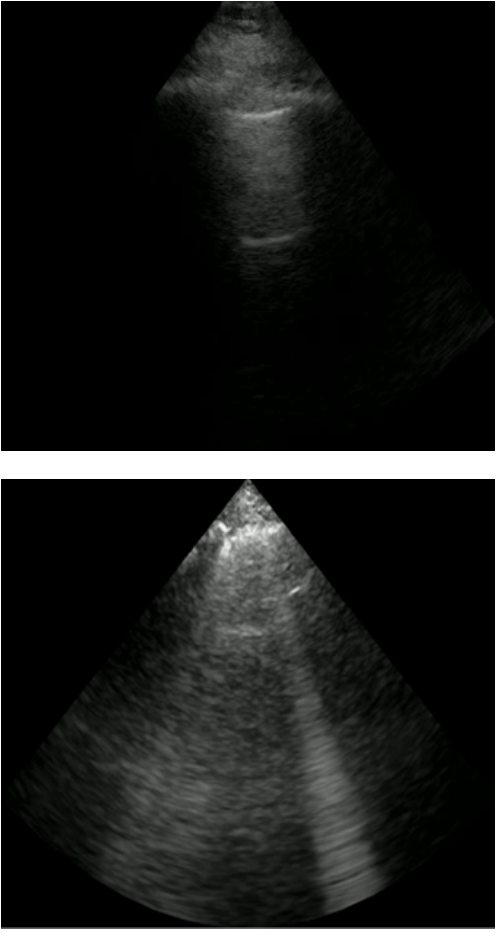}
         \caption{{\tt AB}: A-lines (top), B-lines (bottom) \newline}
         \label{fig:nc-tree}
     \end{subfigure}
    \hfill
     \begin{subfigure}[b]{0.31\linewidth}
         \centering
         \includegraphics[width=0.75\textwidth]{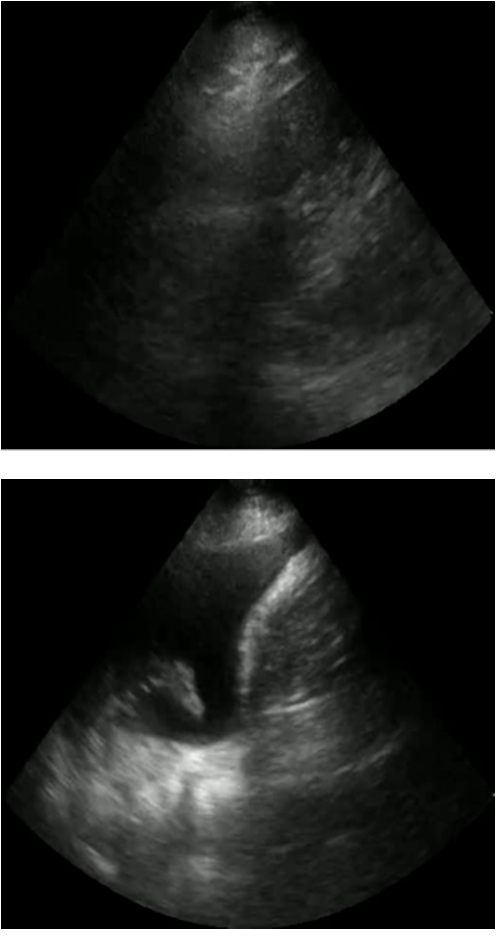}
         \caption{{\tt PE}: No PE (top), PE (bottom) \newline}
         \label{fig:nc-tree}
     \end{subfigure}
    \caption{Examples of each class for each LUS binary classification task: {\tt View} (a), {\tt AB} (b), and {\tt PE} (c).} 
    \label{fig:us-tasks}
\end{figure}

\begin{table*}
    \centering
    \setlength{\tabcolsep}{15pt}
    \begin{tabular}{cp{6cm}cc}
        \toprule
        Task & Description & Negative Class & Positive Class \\
        \midrule
        {\tt View} & Distinguishing the type of LUS view & Parenchymal & Pleural \\
        {\tt A/B} & Normal versus abnormal lung tissue, as indicated by the presence of A-lines and B-lines respectively. Visualized in parenchymal views. & A-lines & B-lines \\
        {\tt PE} & Absence or presence of pleural effusion (PE). Visualized in pleural views. & No effusion & Effusion \\
        \bottomrule
    \end{tabular}
    \caption{A summary of the LUS tasks addressed in this study.}
    \label{tab:lus-tasks}
\end{table*}

\subsection{Data}

Datasets from one local and one external healthcare institution were extracted from a private repository of LUS videos, access to which is permitted via ethics approval granted by Western University (REB 116838). The dataset was previously labelled for the {\tt View}, {\tt AB}, and {\tt PE} tasks by clinicians competent in LUS as a part of prior work~\cite{Arntfield2021automation,Vanberlo2022enhancing}.
The labelled portion of the local dataset was split by patient identifier into a training set ($70\%$), validation set ($15\%$), and test set ($15\%$), and the external dataset was reserved for testing only.
Local videos with no labels were used only during self-supervised pretraining. 
Table~\ref{tab:dataset-description} details the cardinalities and class distribution of these datasets.
Regions peripheral to the US beam were expunged of extraneous visual artefacts, and the images were cropped to the boundaries of the beam. All images were downsampled to $128 \times 128$ pixels.

\begin{table*}[]
    \centering
    \setlength{\tabcolsep}{10pt}
    \begin{tabular}{cccccc}
    \toprule
    & \multicolumn{4}{c}{Local} & External \\
    \cmidrule(r{2pt}){2-5} \cmidrule(l{2pt}){6-6}
    & Unlabelled & Train & Validation & Test & Test \\
    \midrule
    Videos & 9993 & 3545 & 753 & 805 & 374 \\
    Patients & 4919 & 975 & 210 & 210 & 53\\
    Images & \num{2192361} & \num{757492} & \num{161302} & \num{157797} & 45317 \\
    {\tt View} & - & \num{520128}\,/\,\num{237364} & \num{113436}\,/\,\num{47866} & \num{109684}\,/\,\num{48113} & \num{33483}\,/\,\num{11834}\\
    {\tt A/B} & - & \num{195000}\,/\,\num{88361} & \num{43980}\,/\,\num{17269} & \num{43378}\,/\,\num{17719} & \num{12173}\,/\,\num{12078}\\
    {\tt PE} & - & \num{110165}\,/\,\num{40119} & \num{27624}\,/\,\num{8236} & \num{21081}\,/\,\num{9838} & \num{6830}\,/\,\num{2641}\\
    \bottomrule
    \end{tabular}
    \caption{Breakdown of the institutional US datasets used in this study. 
    For each LUS binary classification task, $x\,/\,y$ indicates the number of negative and positive examples respectively.}
    \label{tab:dataset-description}
\end{table*}

We also evaluated the effectiveness of SimCLR-pretrained weights on the public COVIDxUS dataset~\cite{COVIDxUS2021}, splitting it by video identifier into a training, validation, and test set. Although patient identifiers were not available for every video, care was taken to ensure that multiple videos from the same patient identifier were contained in the same set. COVIDxUS contains $243$ LUS videos from a variety of manufacturers and clinical sources with labels for $4$ classes: normal lung, COVID-19, other pneumonia, and other pathologies. 

\subsection{Self-Supervised Pretraining}

\begin{figure}[h]
    \centering
    \includegraphics[width=\linewidth]{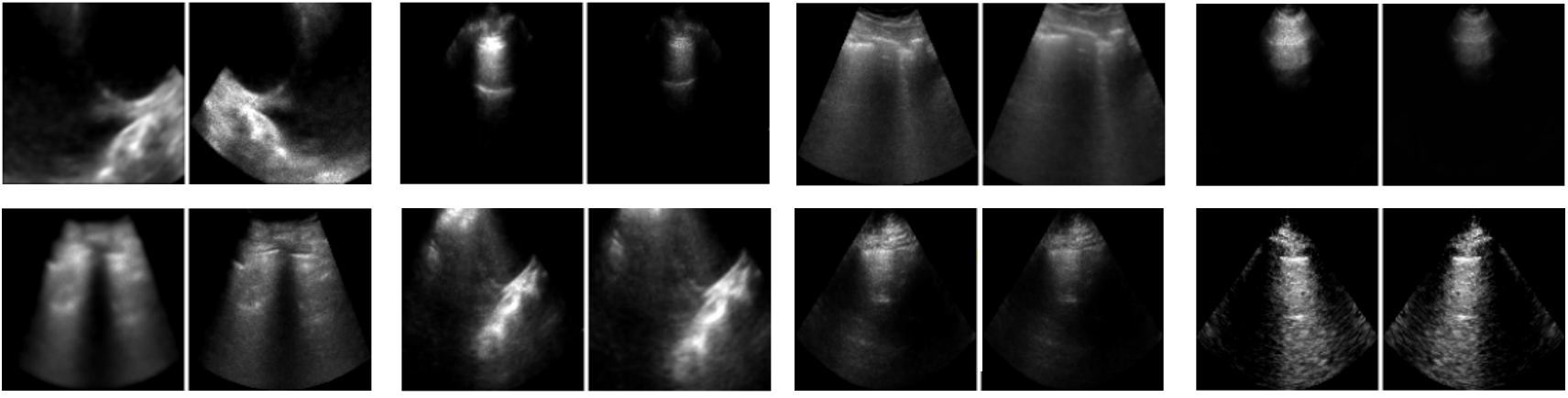}
    \caption{Augmented views of B-mode images, comprising positive pairs for self-supervised pretraining.}
    \label{fig:augmentations}
\end{figure}

Three Joint Embedding SSL methods were trialled to produce pretrained models for each LUS tasks: SimCLR (with $\tau=0.1$)~\cite{chen2020simple}, Barlow Twins (with $\gamma=0.005$)~\cite{zbontar2021barlow}, and VICReg (with $\gamma=25$, $\mu=25$, $\nu=1$)~\cite{bardes2022vicreg}. 
As was done in the original studies, positive pairs were produced by distorting images by applying stochastic data augmentations sampled from a family of transformations.
Figure~\ref{fig:augmentations} provides examples of augmented view of B-mode images from the local dataset.
Below is the list of transformations, where $P$ indicates the probability of that transformation being applied:
\begin{enumerate}
    \item Random crop of $c \sim \mathcal{U}(0.5, 1.0)$ of the image's area. ($P=0.8$).
    \item Horizontal flip. ($P=0.5$)
    \item Multiplicative Gaussian noise, with SD \newline $\sigma \sim \mathcal{U}(0.0, 0.1)$. ($P=0.5$).
    \item Brightness adjustment by $c \sim \mathcal{U}(0.5, 1.5)$. ($P=0.7$).
    \item Contrast adjustment by  $c \sim \mathcal{U}(0.6, 1.0)$. ($P=0.7$). With probability $0.5$, this occurs before brightness adjustment.
\end{enumerate}
Feature extractors were pretrained for $15$ epochs using the union of the unlabelled and training images. 
The MobileNetV3~\cite{howard2019searching} architecture, initialized with ImageNet-pretrained weights, was employed for all pretraining.
The same pretrained feature extractors were used to initialize all downstream LUS tasks.

\subsection{Evaluation Protocol}
\label{subsec:evaluation-protocol}

We compared pretrained models with fully supervised models initialized with ImageNet-pretrained weights.
The following experiments were conducted to determine pretrained models' effectiveness at learning the LUS tasks.
\begin{itemize}
    \item \textbf{Linear classification} (LC): The weights of the feature extractor were held constant, and a linear classifier was trained using its outputted features.
    \item \textbf{Fine-tuning} (FT): The weights of the feature extractor and a linear head were both trained.
    \item \textbf{Nonlinear classification} (NC): The weights of the feature extractor were held constant and a nonlinear head was trained on the features. The head consisted of a multilayer perceptron with a single hidden layer of $32$ nodes with ReLU activation.
\end{itemize}
Figure~\ref{fig:overview} (3a \& 3b) illustrates how FT and NC each implement hierarchical LUS interpretation for the tasks of interest.
In all trials, the initial learning rates for the feature extractor and head were \SI{1e-5} and \SI{1e-4} respectively; they were multiplied by a factor of $e^{-0.02}$ each epoch. 
Models were trained for $10$ epochs to minimize binary cross-entropy loss, and the weights resulting in the lowest validation loss were retained. 
We assessed model performance by determining the area under the receiver operating curve (AUC) on the local and external test sets.
All experiments were conducted using a system with an Intel i9-10900K CPU at \SI{3.7}{GHz} and a Nvidia GeForce RTX 3090 GPU.


\section{Results}

\begin{table*}[h!]
    \setlength{\tabcolsep}{15pt}
    \centering
    \begin{tabular}{llcccccc}
        \toprule
         {\normalsize Task} & {\normalsize  Pretraining} & \multicolumn{3}{c}{{\normalsize Local test set}} & \multicolumn{3}{c}{{\normalsize External test set}} \\
          &  & LC & FT & NC & LC & FT & NC \\
         \midrule
         \multirow{4}{*}{{\large \tt View}} & SimCLR & 0.966 & 0.982 & 0.976 & 0.887 & 0.922 & 0.910 \\ 
          & Barlow Twins & 0.959 & 0.978 & 0.973 & 0.889 & 0.920 & 0.908 \\ 
          & VICReg & 0.960 & 0.980 & 0.975 & 0.894 & 0.917 & 0.909 \\ 
          & None & 0.951 & 0.978 & 0.976 & 0.908 & 0.925 & 0.903 \\ 
         \midrule
         \multirow{4}{*}{{\large \tt AB}} & SimCLR & 0.949 & 0.976 & 0.969 & 0.890 & 0.896 & 0.898\\ 
          & Barlow Twins & 0.940 & 0.958 & 0.948 & 0.903 & 0.886 & 0.894 \\ 
          & VICReg & 0.934 & 0.957 & 0.940 & 0.880 & 0.904 & 0.902 \\ 
          & None & 0.923 & 0.945 & 0.939 & 0.820 & 0.833 & 0.856 \\ 
         \midrule
         \multirow{4}{*}{{\large \tt PE}} & SimCLR & 0.832 & 0.925 & 0.894 & 0.929 & 0.941 & 0.928 \\ 
          & Barlow Twins & 0.855 & 0.906 & 0.893 & 0.845 & 0.940 & 0.923 \\ 
          & VICReg & 0.849 & 0.935 & 0.871 & 0.878 & 0.936 & 0.921 \\ 
          & None & 0.857 & 0.867 & 0.946 & 0.854 & 0.821 & 0.972 \\ 
         \midrule
         \multirow{4}{*}{{\large Mean}} & SimCLR & 0.914 & \bfseries{0.961} & 0.946 & \bfseries{0.902} & \bfseries{0.919} & \bfseries{0.912} \\
          & Barlow Twins & \bfseries{0.917} & 0.947 & 0.937 & 0.879 & 0.915 & 0.908 \\ 
          & VICReg & 0.913 & 0.957 & 0.928 & 0.884 & 0.919 & 0.911 \\ 
          & None & 0.909 & 0.929 & \bfseries{0.954} & 0.860 & 0.858 & 0.909 \\ 
         \bottomrule
    \end{tabular}
    \caption{AUC evaluated on the local and external test sets for the linear classification (LC), fine-tuning (FT), and nonlinear classification (NC) experiments. Results are presented for each of the {\tt View}, {\tt AB}, and {\tt PE} tasks. The bottom row gives the geometric mean across tasks, with bold typeface indicating the best-performing pretraining strategy.}
    \label{tab:test-set-results}
\end{table*}

\subsection{Test Performance}
\label{subsec:test-performance}

Feature extractors were pretrained using SimCLR~\cite{chen2020simple}, Barlow Twins~\cite{zbontar2021barlow}, and VICReg~\cite{bardes2022vicreg}. 
Pretrained models were then fine-tuned for each of the three experiments outlined in Section~\ref{subsec:evaluation-protocol}. 
Table~\ref{tab:test-set-results} details the results on the local and external test sets for each of the experiments described in Section~\ref{subsec:evaluation-protocol}.
Area under the receeiver operating curve (AUC) was designated as the primary evaluation metric, but other classification metrics are reported in Appendix~\ref{apx:classification-metrics}
In the case of linear evaluation, self-supervised pretraining resulted in greater performance on local unseen data. 
Fine-tuned models and MLPs generally achieved greater local test performance, with a notable exception occurring in NC for the {\tt PE} task on the local test set. 

\begin{figure*}[h!]
    \centering
    \includegraphics[width=\textwidth]{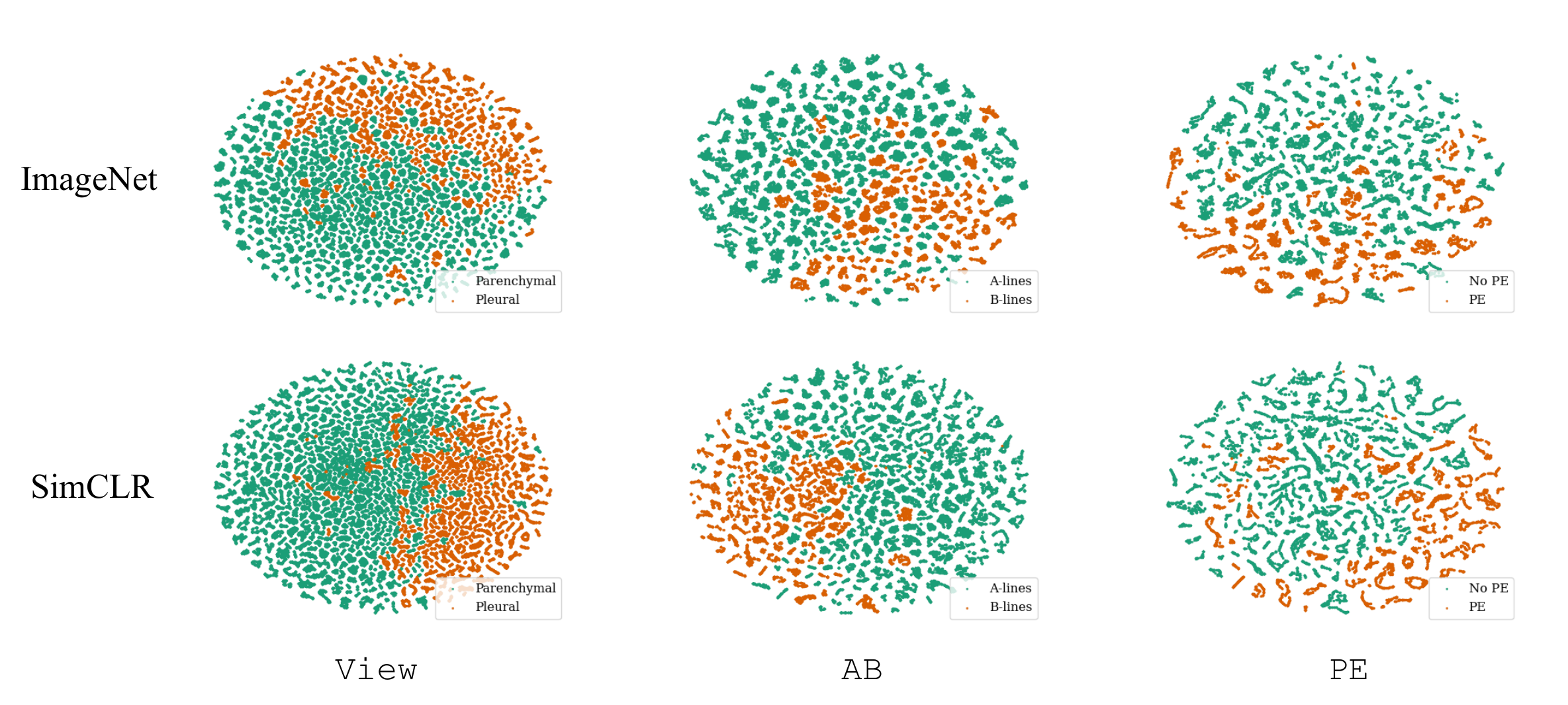}
    \caption{A comparison of t-SNE projections of features for the examples in the local test set outputted by a feature extractor initialized with ImageNet-pretrained weights before SimCLR pretraining (top) and after SimCLR pretraining (bottom). Bold typeface indicates the best-performing pretraining strategy and dataset combination.}
    \label{fig:tsne}
\end{figure*}

Seeking to better understand these results, we visualized two-dimensional t-SNE~\cite{Vandermaaten2008} projections of the features outputted by a ImageNet-pretrained and SimCLR-pretrained feature extractors.
As can be seen in Figure~\ref{fig:tsne}, the projections for {\tt PE} were not well-separated, even after pretraining with SimCLR.
In contrast, the projections suggest that self-supervised pretraining improved the separability of the data for the {\tt AB} task, which is reflected in the ubiquitously stronger performance of the SimCLR-pretrained model.
The difference in performance after self-supervised pretraining was less clear for {\tt View}, which may be because there were significantly more labelled examples available for {\tt View} (see Table~\ref{tab:dataset-description}).
Moreover, the t-SNE projections for {\tt View} exhibited separability before and after SimCLR pretraining. 
As conveyed in Table~\ref{tab:dataset-description}, similar performance trends emerged when evaluating the fine-tuned models on the external test set. 

\begin{table}
    \setlength{\tabcolsep}{7pt}
    \centering
    \begin{tabular}{cccc}
        \toprule
        Experiment & SimCLR  & SimCLR  & Supervised \\
        & [local] & [COVIDxUS] & [ImageNet] \\
        \midrule
        FT & 0.627 & \textbf{0.709} & 0.623 \\
        NC & 0.621 & \textbf{0.751} & 0.685 \\
        \bottomrule
    \end{tabular}
    \caption{Mean class-wise AUC on the COVIDxUS test set for FT and NC.}
    \label{tab:covidx-us}
\end{table}

To promote experimental replicability, we investigated the effect of self-supervised pretraining on COVIDxUS, a public LUS dataset. 
As shown in Table~\ref{tab:covidx-us}, pretraining with SimCLR on the COVIDxUS training set resulted in better mean class-wise test AUC than initialization with ImageNet-pretrained weights.
To explore transferability of pretrained weights, we conducted a separate training run using weights pretrained using SimCLR on the local LUS dataset. 
Note that, although COVIDxUS contains less than a tenth of the number of videos in the local training set alone, it was amalgamated from a variety of institutions and device manufacturers, 
The results highlight the importance of pretraining on a similar distribution, as pretrained weights on the local LUS dataset performed comparably to supervised ImageNet pretraining.

\subsection{Label Efficiency}

\begin{figure*}[h!]
  \centering
  \begin{subfigure}{0.32\linewidth}
    \includegraphics[width=\linewidth]{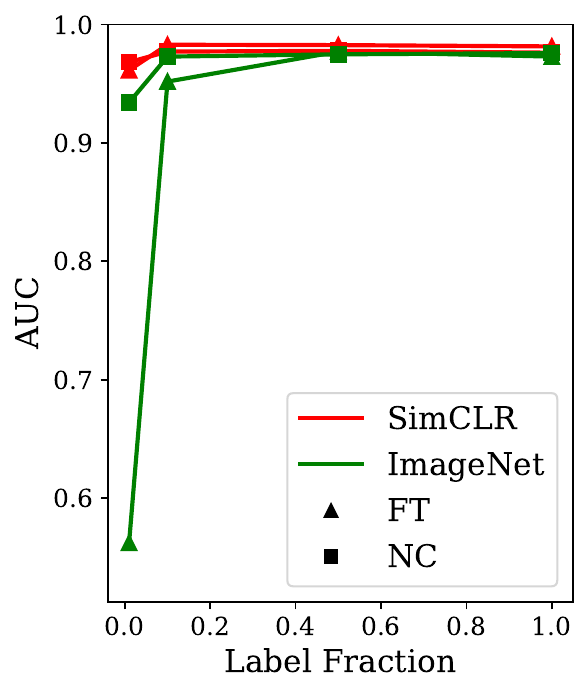}
    \caption{{\tt View}}
    \label{subfig:view-label-eff}
  \end{subfigure}
  \hfill
  \begin{subfigure}{0.32\linewidth}
    \includegraphics[width=\linewidth]{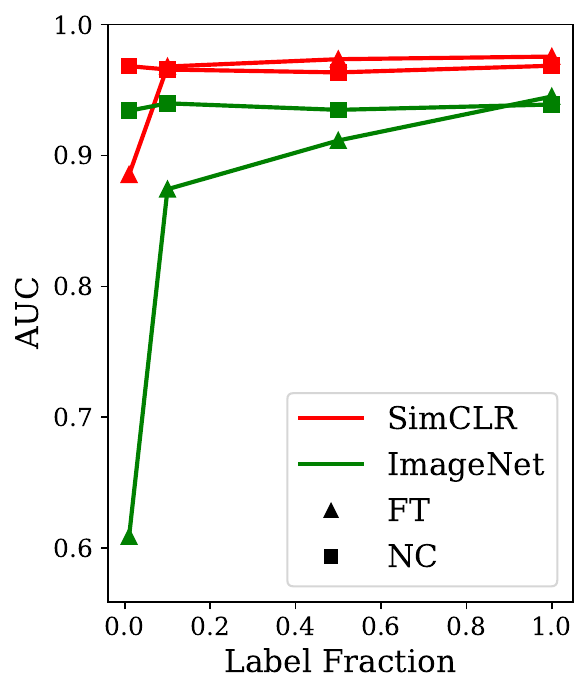}
    \caption{{\tt AB}}
    \label{subfig:ab-label-eff}
  \end{subfigure}
  \hfill
  \begin{subfigure}{0.32\linewidth}
    \includegraphics[width=\linewidth]{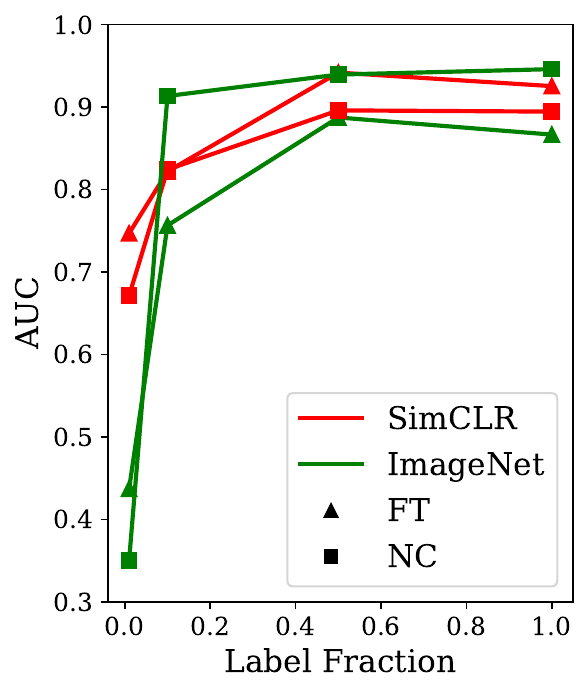}
    \caption{{\tt PE}}
    \label{subfig:pe-label-eff}
  \end{subfigure}
  \caption{Local test AUC for supervised models initialized with ImageNet-pretrained weights and SimCLR-pretrained weights. Results are provided for the fine-tuning (FT) and nonlinear classification (NC) experiments training on various fractions of the labelled dataset.}
  \label{fig:label-efficiency}
\end{figure*}

FT and NC were repeated for ImageNet-pretrained and SimCLR initialization using $1\%$, $10\%$, and $50\%$ of the training set to evaluate the label efficiency of self-supervised pretrained models. 
As depicted in Figure~\ref{fig:label-efficiency}, self-supervision, in most cases, improved performance on the local test set. 
Moreover, the performance gain realized with SimCLR pretraining is largest when training with $1\%$ of the labels. 
A notable exception is the leading performance of the ImageNet-pretrained model with NC for {\tt PE}, which we suspect is due to the greatly reduced number of examples in the test set. 
Again, Figure~\ref{fig:tsne} indicates that the features corresponding to the labelled PE examples are not well-separated by class.

\subsection{Inference Efficiency}
Real-time device inference could be done by reusing the output of a single feature extractor as input to multiple lightweight classifiers.
We compared the inference time of two serial fine-tuned CNNs (Figure~\ref{fig:overview}, 3a) against one feature extractor and two subsequent MLP classifiers (Figure~\ref{fig:overview}, 3b), reflecting the decision tree that results from connecting the {\tt View}, {\tt AB}, and {\tt PE} tasks in the LUS interpretation workflow. 
After serially conducting $1000$ predictions, the former took an average of \SI{0.116}{s} (SD \SI{0.003}{s}), while the latter an average of \SI{0.059}{s} (SD \SI{0.001}{s}), underlining the runtime advantage of multi-task inference with a shared feature extractor.
With each feature extractor and MLP requiring \SI{3.7e7} and \SI{3.7e4} floating point operations respectively, reusing the output of a single feature extractor as input to multiple task-specific MLPs would save considerable computational resources.
The LUS diagnostic tree depicted in Figure~\ref{fig:overview} would require approximately half the floating point operations if each node was a lightweight MLP instead of an entire CNN.
Future work should focus on improving the applicability of representations from frozen self-supervised pretrained models for multiple ultrasound classification tasks.

\section{Conclusion}

In this study, joint embedding SSL methods were observed to improve the performance of classifiers on a variety of LUS tasks, particularly when a small fraction of labels were employed.
Fine-tuning self-supervised pretrained models for each task consistently yielded the greatest performance gains for each task, with SimCLR-pretrained models improving across-tasks average AUC improvement of $0.032$ and $0.061$ on local and external test sets respectively.
When holding the weights of pretrained feature extractors constant, linear classifiers trained on representations from self-supervised models consistently achieved greater across-task average AUC on local and external test data.
MLPs trained on features outputted by self-supervised pretrained models did not outperform fully supervised models on all tasks; however, low-dimensional projections of features provided qualitative evidence that the features were well-separated with respect to two of the three tasks studied.
Given the greatly reduced inference time for multi-task LUS interpretation when reusing features from a single pretrained feature extractor, there would be great merit in future work that improves the quality of pretrained feature extractors and the separability of their outputs with respect to multiple tasks.
As such, future studies could systematically ascertain the effect of US-specific data augmentations in joint embedding methods and explore sample weights for SSL objectives that exploit temporal proximity in B-mode videos.

\section*{Appendix}

\begin{appendix}

\section{Classification Metrics}
\label{apx:classification-metrics}

Tables~\ref{tab:test-set-results-extra-local}~and~\ref{tab:test-set-results-extra-ext} provides additional classification metrics from the evaluation on the local and external test sets respectively (described in Section~\ref{subsec:test-performance}). 

\balance

\begin{table*}[h!]
    \setlength{\tabcolsep}{8pt}
    \centering
    \begin{tabular}{llccccccccc}
        \toprule
         {\normalsize Task} & {\normalsize  Pretraining} & \multicolumn{3}{c}{{\normalsize Precision}} & \multicolumn{3}{c}{{\normalsize Recall}} & 
         \multicolumn{3}{c}{{\normalsize Specificity}} \\
          &  & LC & FT & NC & LC & FT & NC & LC & FT & NC \\
         \midrule
         \multirow{4}{*}{{\large \tt View}} & SimCLR & 0.922 & 0.930 & 0.932 & 0.855 & 0.921 & 0.946 &
         0.968 & 0.969 & 0.972 \\ 
          & Barlow Twins & 0.908 & 0.928 & 0.919 & 0.838 & 0.909 & 0.887 & 0.963 & 0.969 & 0.966 \\ 
          & VICReg & 0.896 & 0.929 & 0.877 & 0.231 & 0.903 & 0.877 & 0.958 & 0.970 & 0.964 \\ 
          & None & 0.913 & 0.912 & 0.916 & 0.799 & 0.906 & 0.879 & 0.967 & 0.961 & 0.965 \\ 
         \midrule
         \multirow{4}{*}{{\large \tt AB}} & SimCLR & 0.893 & 0.934 & 0.902 & 0.795 & 0.839 & 0.815 & 0.961 & 0.976 & 0.964 \\ 
          & Barlow Twins & 0.812 & 0.872 & 0.857 & 0.799 & 0.818 & 0.793 & 0.925 & 0.951 & 0.946 \\ 
          & VICReg & 0.827 & 0.859 & 0.825 & 0.792 & 0.819 & 0.774 & 0.932 & 0.945 & 0.933 \\ 
          & None & 0.852 & 0.778 & 0.885 & 0.677 & 0.861 & 0.760 & 0.952 & 0.899 & 0.960 \\ 
         \midrule
         \multirow{4}{*}{{\large \tt PE}} & SimCLR & 0.824 & 0.908 & 0.848 & 0.593 & 0.680 & 0.662 & 0.941 & 0.968 & 0.945 \\ 
          & Barlow Twins & 0.801 & 0.899 & 0.841 & 0.556 & 0.699 & 0.650 & 0.935 & 0.963 & 0.943 \\ 
          & VICReg & 0.838 & 0.928 & 0.817 & 0.526 & 0.717 & 0.602 & 0.953 & 0.974 & 0.937 \\ 
          & None & 0.925 & 0.996 & 0.940 & 0.468 & 0.371 & 0.667 & 0.982 & 0.999 & 0.980 \\ 
         \bottomrule
    \end{tabular}
    \caption{Classification metrics calculated based on predictions for the local test set for the LC, FT, and NC experiments.}
    \label{tab:test-set-results-extra-local}
\end{table*}

\begin{table*}[h!]
    \setlength{\tabcolsep}{8pt}
    \centering
    \begin{tabular}{llccccccccc}
        \toprule
         {\normalsize Task} & {\normalsize  Pretraining} & \multicolumn{3}{c}{{\normalsize Precision}} & \multicolumn{3}{c}{{\normalsize Recall}} & 
         \multicolumn{3}{c}{{\normalsize Specificity}} \\
          &  & LC & FT & NC & LC & FT & NC & LC & FT & NC \\
         \midrule
         \multirow{4}{*}{{\large \tt View}} & SimCLR & 0.875 & 0.877 & 0.889 & 0.737 & 0.791 & 0.762 & 0.963 & 0.961 & 0.966 \\ 
          & Barlow Twins & 0.892 & 0.931 & 0.930 & 0.725 & 0.769 & 0.724 & 0.969 & 0.980 & 0.981 \\ 
          & VICReg & 0.946 & 0.918 & 0.926 & 0.672 & 0.771 & 0.720 & 0.986 & 0.976 & 0.980 \\ 
          & None & 0.783 & 0.846 & 0.811 & 0.664 & 0.769 & 0.737 & 0.935 & 0.951 & 0.939 \\ 
         \midrule
         \multirow{4}{*}{{\large \tt AB}} & SimCLR & 0.906 & 0.866 & 0.876 & 0.676 & 0.736 & 0.729 & 0.929 & 0.885 & 0.896 \\  
          & Barlow Twins & 0.931 & 0.904 & 0.901 & 0.704 & 0.742 & 0.726 & 0.948 & 0.921 & 0.920 \\ 
          & VICReg & 0.935 & 0.921 & 0.928 & 0.684 & 0.726 & 0.698  & 0.952 & 0.937 & 0.945 \\ 
          & None & 0.810 & 0.730 & 0.847 & 0.612 & 0.830 & 0.665 & 0.855 & 0.691 & 0.880 \\ 
         \midrule
         \multirow{4}{*}{{\large \tt PE}} & SimCLR & 0.979 & 0.995 & 0.967 & 0.663 & 0.643 & 0.627 & 0.993  & 0.998 & 0.989\\ 
          & Barlow Twins & 0.936 & 0.996 & 0.984 & 0.588 & 0.605 & 0.615 & 0.979 & 0.999 & 0.995 \\ 
          & VICReg & 0.876 & 0.998 & 0.982 & 0.618 & 0.602 & 0.649 & 0.954 & 0.999 & 0.994 \\ 
          & None & 0.739 & 1.00 & 0.999 & 0.482 & 0.235 & 0.597 & 0.910 & 1.00 & 1.00 \\ 
         \bottomrule
    \end{tabular}
    \caption{Classification metrics calculated based on predictions for the external test set for the LC, FT, and NC experiments.}
    \label{tab:test-set-results-extra-ext}
\end{table*}

\end{appendix}

\bibliographystyle{unsrt}  
\bibliography{references}

\end{document}